\documentclass{article}

\usepackage{PRIMEarxiv}

\usepackage[utf8]{inputenc} 
\usepackage[T1]{fontenc}    
\usepackage{hyperref}       
\usepackage{url}            
\usepackage{booktabs}       
\usepackage{amsfonts}       
\usepackage{nicefrac}       
\usepackage{microtype}      
\usepackage{lipsum}
\usepackage{graphicx}
\usepackage{subcaption}
\usepackage{mwe}
\usepackage{amsmath}
\usepackage{array}
\usepackage{float}
\usepackage{multicol}
\usepackage{csquotes}
\usepackage{caption}
\usepackage{multirow}   

\title{On the balance between the training time and interpretability of neural ODE for time series modelling
\thanks{\textit{\underline{Citation}}: 
\textbf{Authors. Title. Pages.... DOI:000000/11111.}} 
}

\author{
  Yakov Golovanev, Alexander Hvatov \\
  NSS Lab \\
  ITMO University \\
  St Petersburg, Russia\\
  \texttt{alex\_hvatov@itmo.ru} \\
}

\begin{document}
\maketitle

\begin{abstract}
Most machine learning methods are used as a black box for modelling. We may try to extract some knowledge from physics-based training methods, such as neural ODE (ordinary differential equation). Neural ODE has advantages like a possibly higher class of represented functions, the extended interpretability compared to black-box machine learning models, ability to describe both trend and local behaviour. Such advantages are especially critical for time series with complicated trends. However, the known drawback is the high training time compared to the autoregressive models and long-short term memory (LSTM) networks widely used for data-driven time series modelling. Therefore, we should be able to balance interpretability and training time to apply neural ODE in practice.
The paper shows that modern neural ODE cannot be reduced to simpler models for time-series modelling applications. The complexity of neural ODE is compared to or exceeds the conventional time-series modelling tools. The only interpretation that could be extracted is the eigenspace of the operator, which is hard to obtain for a large systems. Spectrum could be extracted using different classical analysis methods that do not have the drawback of extended training time. Consequently, we reduce the neural ODE to a simpler linear form and propose a new view on time-series modelling using combined neural networks and ODE systems approach.
\end{abstract}


\section{Introduction}
    \label{sec:intro}
    
Continuous-time machine learning models have a long history starting from classical ARIMA models to most modern liquid neural networks \cite{liquid-time-constant-networks}. Most of the contemporary methods are not interpretable \cite{lipton2018mythos} in terms of both model behaviour in the \enquote{unknown} valid input case and the model parameters sensitivity (which is denoted as \enquote{Simulatability} in \cite{lipton2018mythos}) and, which is more critical to time-series, we are not able to extract scales from the model like a trend, seasonal component and others (which is denoted as \enquote{Decomposability} in \cite{lipton2018mythos}). 

Conventional AI approaches to time-series modelling are usually interpreted expertly, either using area knowledge \cite{srilakshmi2022improving}, or visualization \cite{shen2020visual}. There are also exist more automated methods such as STAM \cite{gangopadhyay2021spatiotemporal} (spatio-temporal attention mechanism). Such an interpretation is made after the prediction is obtained. The conventional machine learning model alone cannot give any insights into the modelling process. Physical analogies such as physical-informed networks \cite{raissi2019physics}, differential equations discovery \cite{brunton2016discovering,maslyaev2021partial} or neural ODE \cite{chen2018neural} may possibly give the model that could be self-explainable.

Recent research on neural ODEs (ordinary differential equations)~\cite{chen2018neural} is a separate branch that achieved impressive results in several areas, including replacement of residual networks in classification problems, constructing a new class of invertible density models, and proposing a generative time series model. Systems of ODEs have a long history of classical analysis, thus having well-developed tools for modelling quality assessment and interpretation. Consequently, the neural ODE may also be interpreted from a classical ODE analysis point of view.

There are two \enquote{mainstream} analytical points of view on neural ODE. The first direction is using methods of topology represented by \cite{dupont2019augmented} as an example. The topological analysis mainly may be used to interpret classification-like models. Namely, in the scope of the topological approach are the properties of mapping between feature and target spaces. The paper~\cite{dupont2019augmented} reveals some of the limitations of the original model and resolves them by extending this model. The authors showed that the dynamics of neural ODE define homeomorphism that continuously deforms the input space and cannot, for example, tear a connected region apart. To address discovered theoretical limitations, the authors proposed the augmented neural ODEs, which employ additional dimensions to allow learning trajectories that would have self-intersections in the original space.

The second direction is instead a functional analysis point of view. In~\cite{dissecting-neural-odes} authors established a general system-theoretic neural ODE formulation and analyzed its components separately. Authors show that the original neural ODE model from~\cite{chen2018neural} in fact, cannot be considered as a limit of the deep residual networks. Similarly, apart from the theoretical study of existing neural ODE models, authors, starting from the problem of building proper deep limit of residual networks, introduced two novel model variants --- Galerkin and stacked neural ODEs. The functional analysis point of view is more applicable to extending the very deep neural networks such as~\cite{hamiltion-cnn} that can be viewed as various ODE discretizations.

Classical ODE system analysis (e.g. stability and spectrum analysis) seems a more appropriate analysis tool for time-series applications. However, it appears more rarely. In~\cite{neural-ode-review} authors show the relation between training a continuous-time neural network and solving an optimal control problem. In~\cite{latent-ode-for-time-series} authors of the original neural ODE paper consider problems of extrapolation and interpolation of irregularly-sampled time series data. The proposed time series model is based on a model from~\cite{chen2018neural}, extending it by using an ODE encoder with a recurrent neural network (RNN). The second application is also based on the results from~\cite{chen2018neural} and is related to the proposed density modelling approach. In~\cite{ffjord}, authors used Hutchinson's trace estimator for estimating the log-density in linear time, allowing unrestricted architectures for the derivative of the hidden state.

In practical applications, we are primarily interested in balancing the modelling time and either the quality of result or interpretation, i.e. what we obtain in exchange for time. Paper~\cite{learning-easy-to-solve} addresses the critical limitation of modern neural ODE  models --- computational complexity --- by enforcing the generated ODEs to be easy to solve. The authors introduced a differentiable surrogate using higher-order derivatives of solution trajectories to achieve this. Thus, we should be able to extract properties or generalize results compared to the conventionally used machine learning models for time-series modelling. The conventional tools are fast but lack interpretability and generalization that could be replaced by neural ODE.

In this work, we consider the problem of time series modelling using neural ODE from a time-series practical modelling point of view. Previous papers significantly extend the original model using the conventional functional analysis and machine learning methods, despite detailed research on different properties of neural ODEs. Many questions remain unanswered regarding these models' theoretical and practical properties.
Two of the arisen questions with a brief explanation of the related problems are discussed below.

    \begin{itemize}
        \item \textbf{In which case neural ODE models should be used instead of classical time series models?} Authors of~\cite{latent-ode-for-time-series} showed that when data is sparse or irregularly sampled, these models perform better than RNN networks. Apart from this, unlike standard RNN networks and other time series models, hidden states of neural ODE models correspond to the coefficients of systems of differential equations, which increases the overall interpretability of the models, which is crucial in cases where black-box models cannot be trusted. It is necessary to compare its behaviour with various classical time series models on various datasets to determine the limits of applicability of this class of models.
        \item \textbf{What is the maximal information we can obtain to analyze the neural ODE using the classical analysis methods?} Eigenspace analysis is a powerful tool to assess the stability of the equation. Neural ODE produces, in most general form non-linear equation system. Whereas linear systems are analyzed similarly, each non-linear system requires either a separate approach to find spectrum (in this case, spectrum consist of three parts: discrete, continuous and residual) or linearization. For matrix eigenspace numerical analysis (which is essentially the same as linear ODE system spectrum analysis) for higher dimensions, the computation of the eigenvalues is a well-known ill-posed problem. It requires an individual approach every time it appears. Thus, this question also refers to the amount of the parameters we use to build a neural ODE model for practical time-series application.
    \end{itemize}

The first question could be expanded onto balance between interpretability, prediction quality and training/inference time question for the machine learning methods for the time series prediction. Answer to this question should also lead to the idea of what theoretical interpretation limit may be obtained with the latent neural ODE approach and how we can extend it further.

The second question is bound to the model's informational capacity. We require as few parameters as possible to interpret the neural ODE model since we cannot work with large ODE systems spectrum. Both questions lead to the classical complexity and interpretability trade-off problem that becomes sharper when considering ODE systems.

In the paper, we stoplight the following points based on the previous analysis:

\begin{itemize}
    \item For practical applications number of neural ODE parameters is large and does not allow for computing the eigenspace of the non-linear system. Spectrum computation is an ill-posed problem for large systems. Thus, we want not only to balance complexity but overall problem posedness.
    \item We may use linear ODE systems (neural ODE without activation functions between layers) as an attempt at ad-hoc linearization of the problem. However, such a statement from an interpretation point of view is easily reduced to the symbolic regression in a solution space, which speeds up the optimization process and does not lose the model's informational capacity.
    \item The interpretable practical work requires other than neural ODE approaches to combine neural networks and ODE systems. Namely, we propose to use the composite models.
\end{itemize}

The paper is organized as follows: Section~\ref{sec:problem_statement} introduces the neural ODE principles. Section~\ref{sec:ML_compare} contains a comparison of basic neural ODE methods with the classical machine learning algorithms for the time series prediction. Section~\ref{sec:node_analysis} is dedicated to the analysis of two possible forms of neural ODE representations from a classical ODE analysis point of view. Section~\ref{sec:discussion} and Section~\ref{sec:conclusion} contain main insights obtained during the research and future work directions.

 \section{Problem statement}
\label{sec:problem_statement}

\paragraph{Classical neural ODE training problem.} A classical neural network consists of a discrete set of transformations on the input vector, resulting in a dense representation that allows solving various problems. This framework is ubiquitous, and a vast number of various transformations can be applied. However, for many practical tasks, the treatment of the output data as the result of a sequence of discrete transformations on input data is inaccurate and unnatural~\cite{chen2018neural}. One noticeable example is the time series data, which are usually continuous. For this reason, the problem of training models, which can adapt to this kind of data, arises. The critical observation is that most of the neural network architectures, such as residual networks~\cite{resnet}, can be viewed as the sequence of transformations in the form~\eqref{eq:nn_transform}.
    
    \begin{equation}
    \label{eq:nn_transform}
        h_{t + 1} = h_{t} + f(h_{t}, \theta)
    \end{equation}
    
    In~\eqref{eq:nn_transform} $h_t \in \mathbb{R}^d$ is the hidden state after $t$-th transformation, and $\theta$ is the network parameter vector. This equation is in fact the Euler's discretization of the following differential equation (with step size equal to $1$):
    
    \begin{equation}
        \label{eqn:neuralode_transform}
        \lim \limits_{\tau  \to 0} \frac{h_{t + \tau} - h_{t}}{\tau}=  \frac{d h(t)}{d t} = f(h(t), t, \theta)
    \end{equation}
    
In~\eqref{eqn:neuralode_transform} $\tau^{-1}$ may be considered, with some limitations mentioned in~\cite{dissecting-neural-odes}, as the density (number) of layers between the moments $t$ and $t+1$. Thus, by solving an equation, one can obtain an \enquote{infinitely deep} version of the residual network\@. The forward pass of the ODE network requires integrating the solution for equation~\eqref{eqn:neuralode_transform} over the selected time domain. On the contrary, performing the backpropagation through such a layer is challenging since the direct differentiation of solver operations requires significant computational efforts. While the idea of training continuous-time neural networks was initially proposed in the 80s of the last century~\cite{backprop}, ~\cite{dynamic}, an important contribution is made by~\cite{chen2018neural}, authors used the adjoint sensitivity method from the late 60s~\cite{pontryagin1962} that allowed computing gradients by solving a second, augmented ODE backwards in time.

\begin{figure}[ht!]
            \centering
            \includegraphics[width=0.45\linewidth]{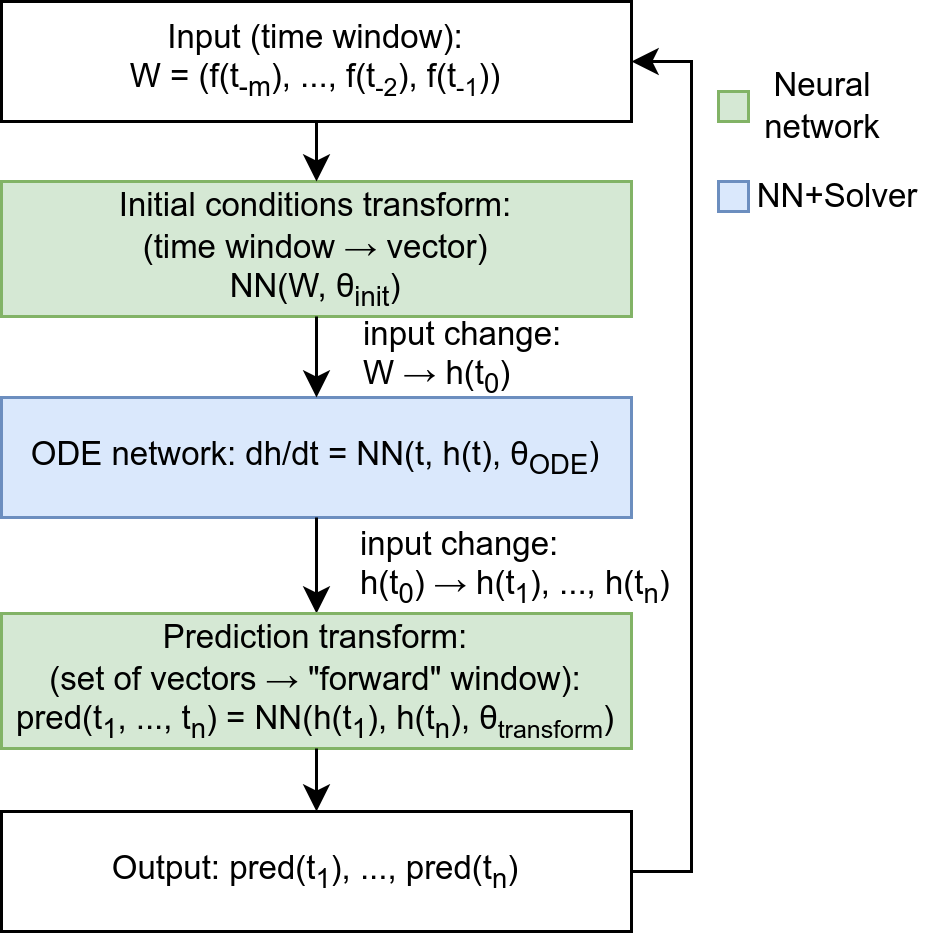}
            \caption{Classical neural ODE workflow.}
            \label{fig:neural_ODE_scheme}
        \end{figure}

\paragraph{Neural ODE interpretation problem.} From a machine learning point of view, neural ODE training for time series prediction consists of three separate models. The first model should transform the given time series window into initial conditions for the differential equation system solver. The second model is the right-hand side of the differential equation system. After that, the final model transforms vectors obtained after system integration back into time series. After the forward computation mode, the prediction is used for backpropagation for tuning the three models' parameters. The resulting scheme of neural ODE workflow is shown in Figure~\ref{fig:neural_ODE_scheme}.

We have to interpret three neural networks within the neural ODE training process. First - the initial condition transform should be interpreted as an a priori guess for decomposed time-series vector start. Theoretically, initial conditions govern the function set from an operator eigenvectors space to form the solution. Since the form of the operator is also a neural network, it is hard to interpret this neural network from an analytical point of view. Empirically we can consider the plot for each vector item and assess its continuity.

The second neural network contains most of the information on the process and could be interpreted from a classical ODE analysis point of view. Since we obtain a non-linear system of ODE (even though it is \enquote{encrypted} by the neural network), we can obtain the continuous and point-wise spectrum numerically. It could help assess the system stability and thus the prediction horizon, i.e. how often we should recompute the initial condition. We could also recompute the system. However, we are interested in preserving the \enquote{physics} of the model (otherwise, we do not have any reasons to use neural ODE out of deep neural network applications).

The third neural network is usually the linear combination of items in the system's time-series vector. We cannot extract much from it apart from its continuity in the correctly trained networks system and the possible connection with the first network.

\paragraph{Spectrum as an ODE systems interpretation.} Whereas interpretation of first and third neural networks are simple, interpretation of non-linear systems may become a complicated task. We show main tools for ODE interpretation systems in Table~\ref{tab:ODE_interpretation}.

\begin{table}[h!]
\centering
\caption{Types of the systems and the interpretation tools.}
\begin{tabular}{ll}
\hline
                \toprule
System type                & Interpretation tools                                                                                                                                            \\ \midrule
Larger non-linear systems  & \begin{tabular}[c]{@{}l@{}}- decomposition to a simpler systems \\ - very rarely interpretable as a whole\end{tabular}                                          \\ \midrule
Simpler non-linear systems & \begin{tabular}[c]{@{}l@{}}- linear part of spectrum\\ - stability\\ - different form of continuous spectrum\\ (separate for every class of systems)\end{tabular} \\ \midrule
Larger linear systems      & \begin{tabular}[c]{@{}l@{}}- ill-conditioned\\ - linear(ized) spectrum\\ - decomposition\end{tabular}                                                            \\\midrule
Linear systems             & - linear(ized) spectrum                                                                                                                                          \\ \midrule
Solutions                  & \begin{tabular}[c]{@{}l@{}}- possible equation restoration\\ - linear(ized) spectrum\end{tabular}                                                                \\ \bottomrule \hline
\end{tabular}
\label{tab:ODE_interpretation}
\end{table}

\begin{figure}[h!]
    \centering
    \includegraphics[width=0.4\linewidth]{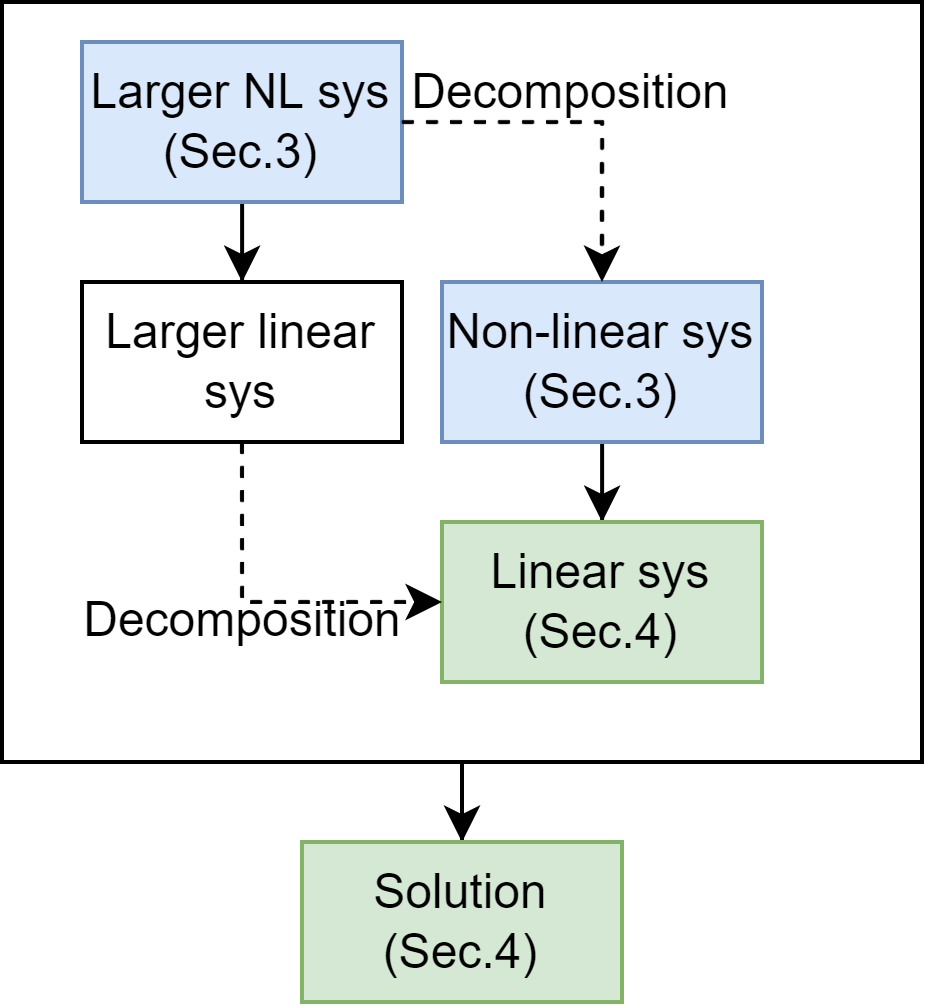}
    \caption{Non-linear ODE systems decomposition scheme.}
    \label{fig:ODE_interp}
\end{figure}

We usually linearize the system to extract spectral characteristics. Linearized spectrum for the significantly non-linear system contain the influence of the high-order decomposition part. We note that overall there is no notion of \enquote{non-linear spectrum}. It may be defined from various points of view, and the choice of the definition depends on an operator class and application. Thus, to analyze the non-linear systems more efficiently, we usually refer to the linearized part of the spectrum. Figure~\ref{fig:ODE_interp} shows how the problems from Table~\ref{tab:ODE_interpretation} are connected.

In Section~\ref{sec:ML_compare} we further discuss how large the non-linear ODE system should be to forecast time series. Since the right-hand side is a neural network, it would be hard to decompose it into smaller systems. Therefore, we aim to obtain as a small network as possible.

In Section~\ref{sec:node_analysis} we discuss if the neural ODE without activation layers provides a linear system faster than more classical tools such as Fourier decomposition. We could use the neural ODE as a system linearization tool if it were possible to obtain a linear system within a reasonable time range.

\section{Neural ODE complexity assessment}
\label{sec:ML_compare}

Differential equations systems, including ODE network step in Figure~\ref{fig:neural_ODE_scheme}, are very natural for time series data. 
We consider the time series forecasting problem on several physics-related datasets to build the model capable of predicting $n$ subsequent observations based on the history of $m$ observations. The models are compared based on \textit{MAE} (mean averaged error) metrics for each dataset with different noise levels. The goal of using the neural ODE model is to minimize the values of these metrics and achieve explainability due to the ODE nature of the considered model. In experiments, we compared the performance of the latent ODE model from~\cite{latent-ode-for-time-series} with several traditional time series models.

As the starting point, we conduct several experiments on different physics-related datasets. We select two time series datasets with non-stationary data to comprehensively study the representative properties of neural ODE models compared to several classical models. For the first experiment, we choose the outdoor air quality dataset~\cite{air-quality-dataset}, selecting ten years of hourly sampled temperature data at a specific location. For the second experiment, we select the Venice water level dataset~\cite{venice-water-dataset} with hourly sampled water level data at a specific point measured hourly from $1983$ to $2015$.

In both experiments, we split the dataset in three parts --- training, validation and test in proportions $0.7$ / $0.2$ / $0.1$. In experiments, the data is scaled to have zero mean and standard deviation equal one.
On selected datasets, we compared five models --- repeater model, which returns its input, fully-connected neural network, ARIMA model, LSTM model, and latent ODE model from~\cite{latent-ode-for-time-series}. 

In these experiments, each model except the repeater has $m=100$ data points as the input, and each model is trained to produce the next $n$ data points. We conducted several experiments, setting $n$ equals $100$, $250$ and $500$ and repeating the training and evaluation $k=5$ times.
For the repeater model, we set $m = n$ in each experiment.
Apart from this, we added noise $\epsilon \sim \mathcal{N}(\mu,\,\sigma^{2})$ to the data, varying $\sigma$ from $0$ to $0.3$ with step of $0.1$.
This results in $2$ (number of datasets) $\times \, 3$ (number of considered values of $n$ ) $\times \, 4$ (number of considered noise levels) $ = 24$ different experiments, each of them repeated $k=5$ times.

\subsection{Description of compared models}
\label{sec:models}

The fully-connected and LSTM models were trained for a $200$ epoch each, with batch size set to $32$.
For simplicity, we used Keras~\cite{keras} implementation of these models.
The fully-connected model consists of one hidden layer with $128$ units and ReLU activation and contains different number of parameters for different values of $n$: $13\,156$ for $n=100$, $32\,506$ for $n=250$, and $64\,756$ for $n=500$.
The LSTM model consists of single LSTM layer with $32$ units and a fully-connected network equivalent to the previously mentioned one on top of it. This results in $21\,476$ parameters for $n=250$, $40\,826$ for $n=250$ and $73\,076$ for $n=500$.
For both models, increasing the number of training epochs or the number of network parameters did not improve the test accuracy.

The seasonal ARIMA model with the period of $24$ was used for both datasets, as in both cases, it corresponds to the natural daily period in data.
This model consists of only four parameters: constant bias, AR and MA terms, and standard deviation for a residual term.
Including additional AR or MA terms and using an integrated or differencing model did not improve the model significantly.

The latent ODE model from~\cite{latent-ode-for-time-series} has the most complex structure over considered models.
For our experiments, we used the same setup that was used for modelling 1D periodic functions in ~\cite{latent-ode-for-time-series}
while increasing the size of the ODE part latent state to $15$.
The resulting model consists of $55\,561$ parameters independent of the value of $n$.
As we will discuss in Section~\ref{sec:node_analysis}, this is much more than is necessary for modelling considered time series using
ODE models, however, decreasing the number of parameters leads the latent ODE model to be unable to reproduce the periodic components in the data, while increasing the number of parameters did not significantly improve the prediction quality or convergence speed.
Moreover, this model required many iterations to converge --- up to $9000$, which causes the slowest training process among the considered models. 

\subsection{Experiments with temperature dataset}
\label{sec:temp_experiment}
   
\begin{figure}[ht!]
            \centering
            \includegraphics[width=0.5\linewidth]{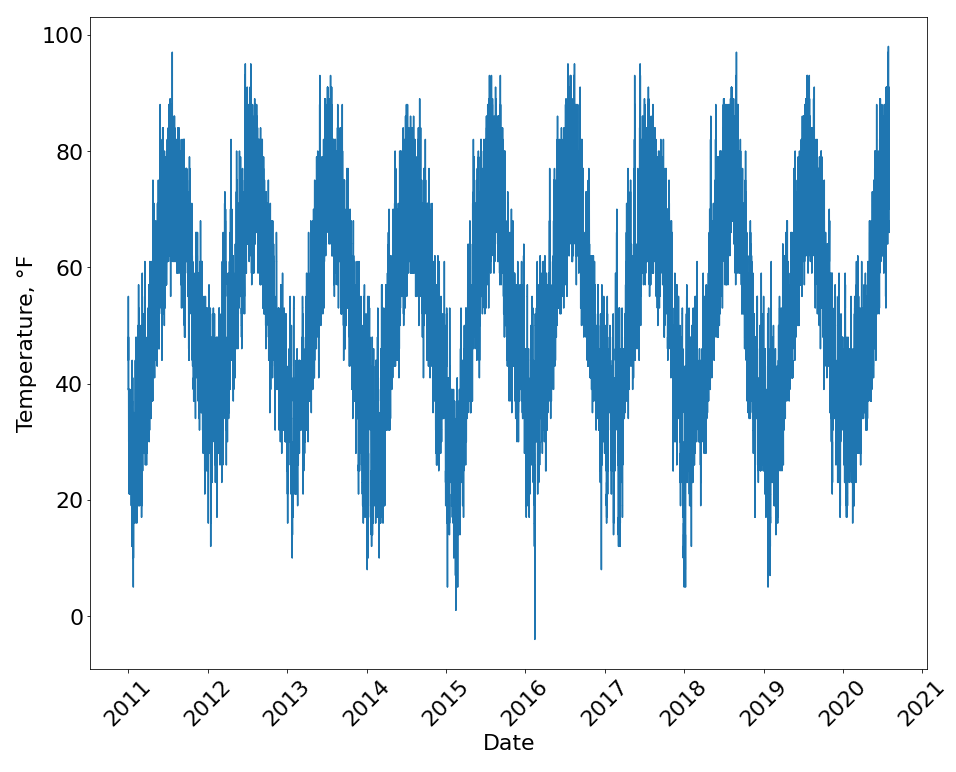}
            \caption{Air temperature for ten years in Barnstable, Massachusetts.}
            \label{data-example-temp}
        \end{figure}

    For this experiment, we use the temperature data for a specific location in Barnstable, Massachusetts, which is shown in Figure~\ref{data-example-temp}.
    
\begin{figure*}
        \centering
        \begin{subfigure}[b]{0.495\textwidth}
            \centering
            \includegraphics[width=\textwidth]{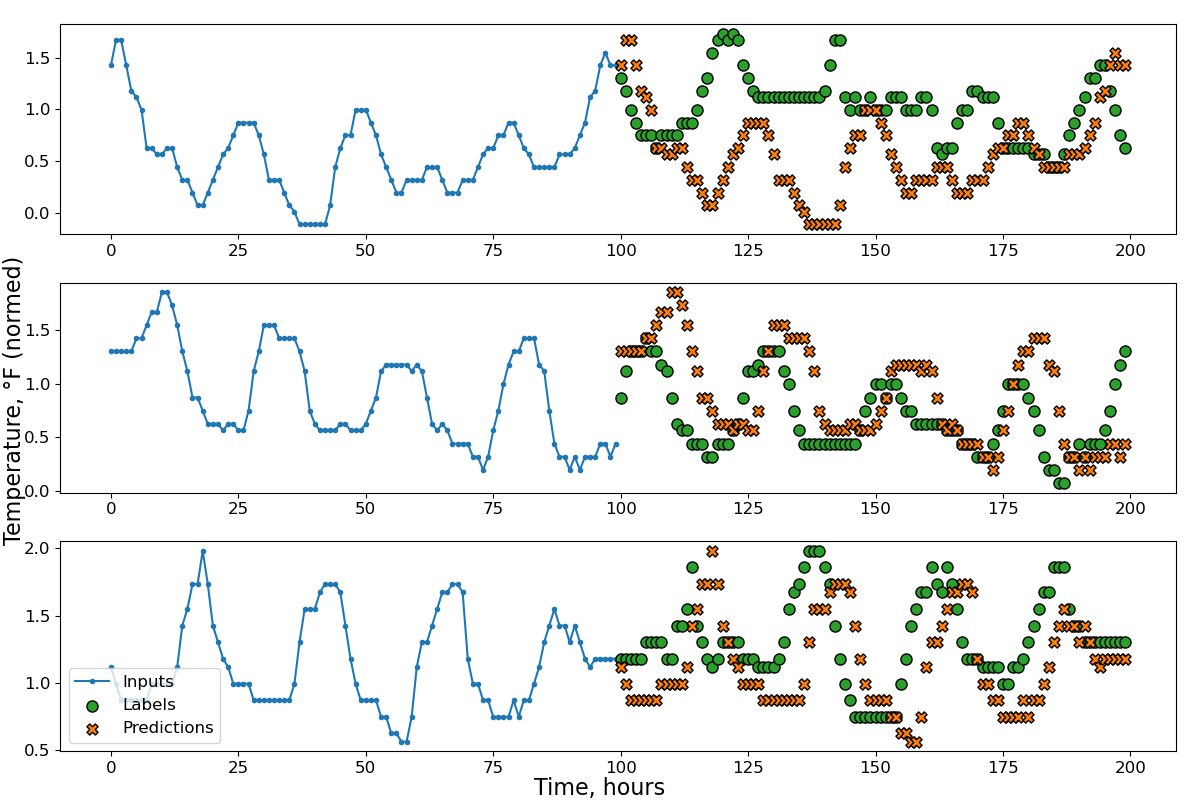}
            \caption[Network2]%
            {\small{Repeater model}}    
            \label{fig:temperature_picked_prediction_repeater}
        \end{subfigure}
        \hfill
        \begin{subfigure}[b]{0.495\textwidth}  
            \centering 
            \includegraphics[width=\textwidth]{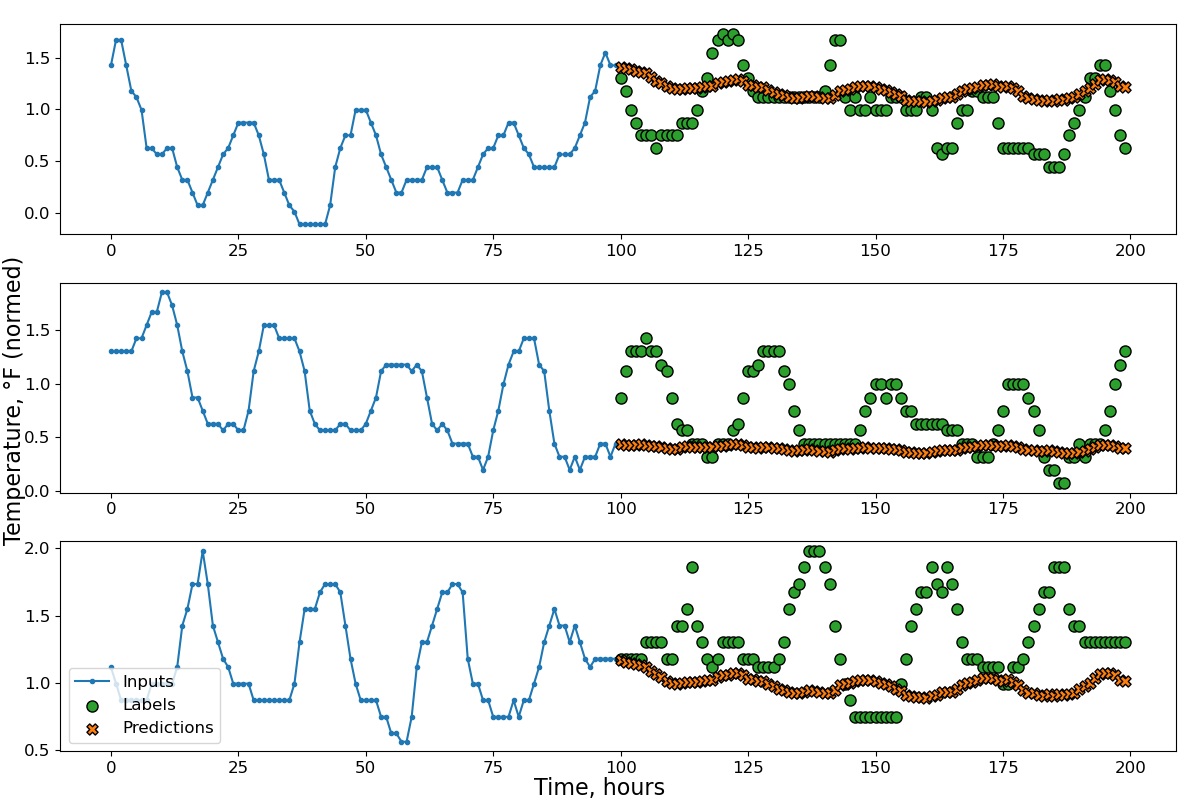}
            \caption[]%
            {\small{Dense neural network}}
            \label{fig:temperature_picked_prediction_dense}
        \end{subfigure}
        \begin{subfigure}[b]{0.495\textwidth}   
            \centering 
            \includegraphics[width=\textwidth]{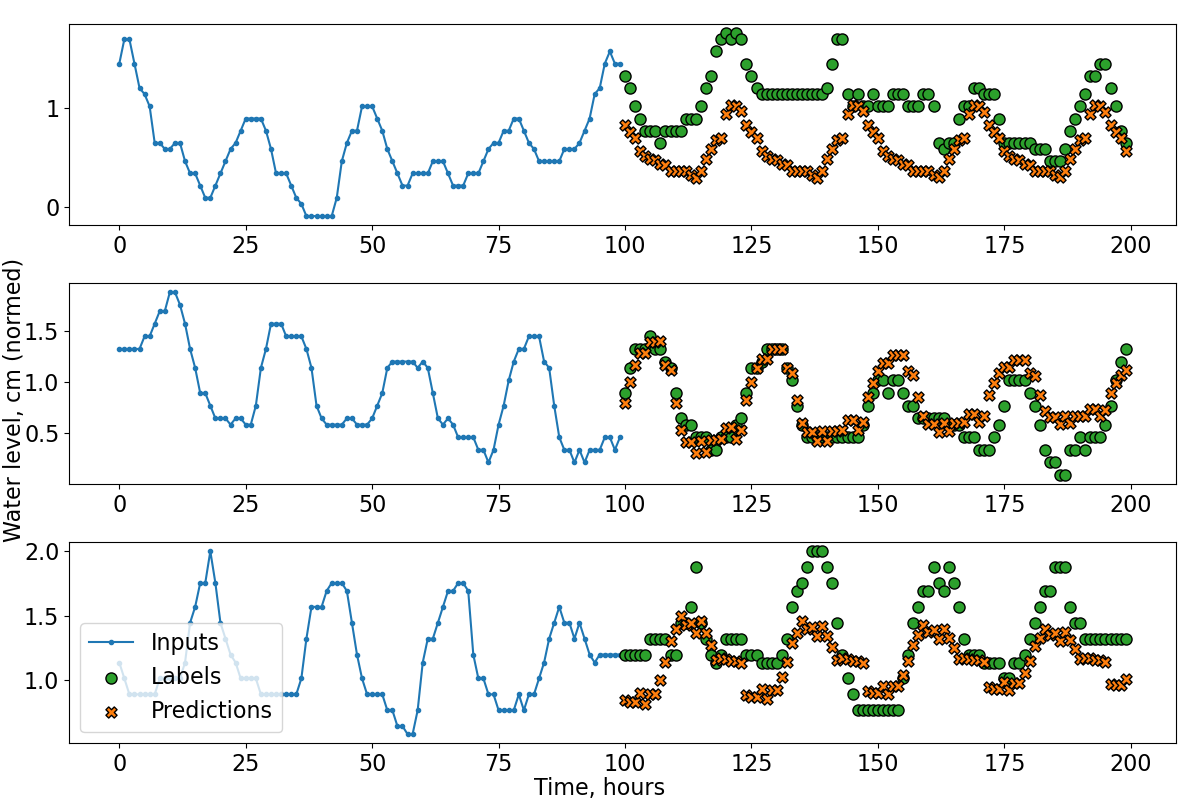}
            \caption[]%
            {\small{Autoregressive model}}
            \label{fig:temperature_picked_prediction_autoreg}
        \end{subfigure}
        \hfill
        \begin{subfigure}[b]{0.495\textwidth}   
            \centering 
            \includegraphics[width=\textwidth]{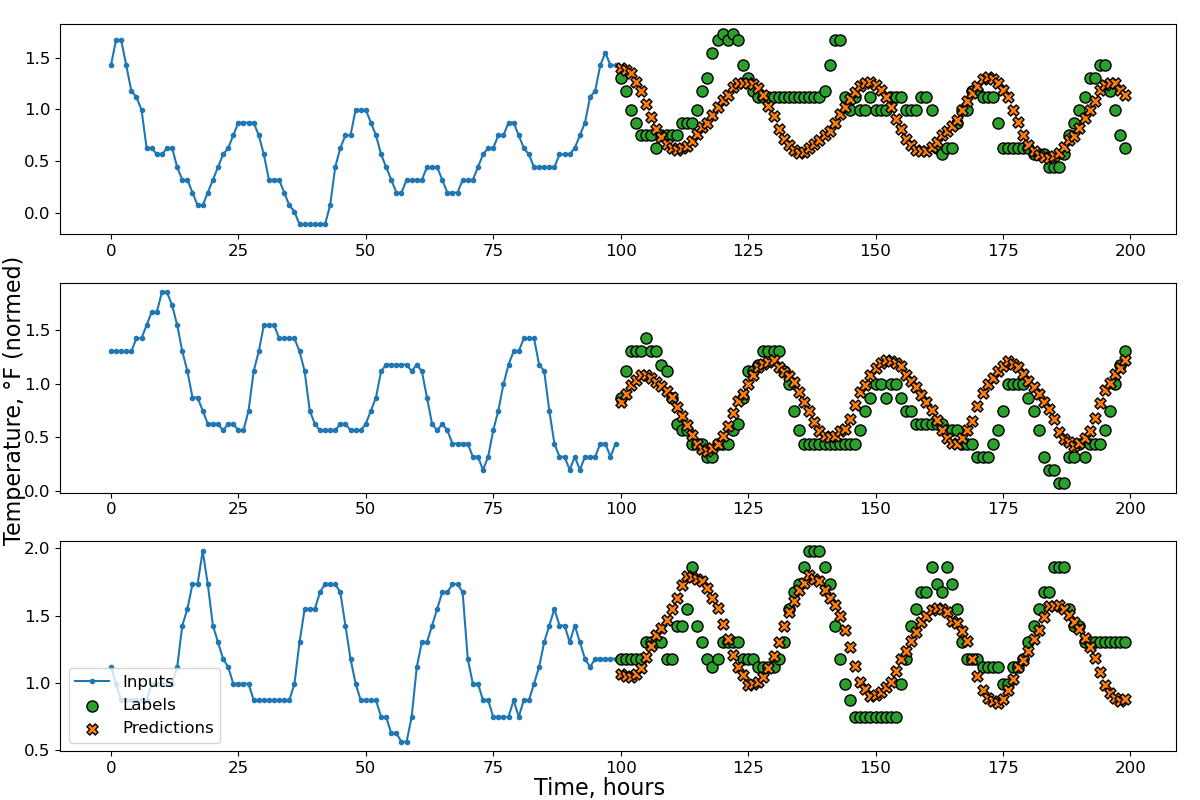}
            \caption[]%
            {\small{LSTM model}}
            \label{fig:temperature_picked_prediction_lstm}
        \end{subfigure}
        \caption[ Picked predictions of a temperature dataset for different models. ]
        {\small Picked predictions of a temperature dataset for different models.} 
        \label{fig:temperature_picked_prediction}
    \end{figure*}

\begin{figure}[ht!]
            \centering
            \includegraphics[width=0.5\linewidth]{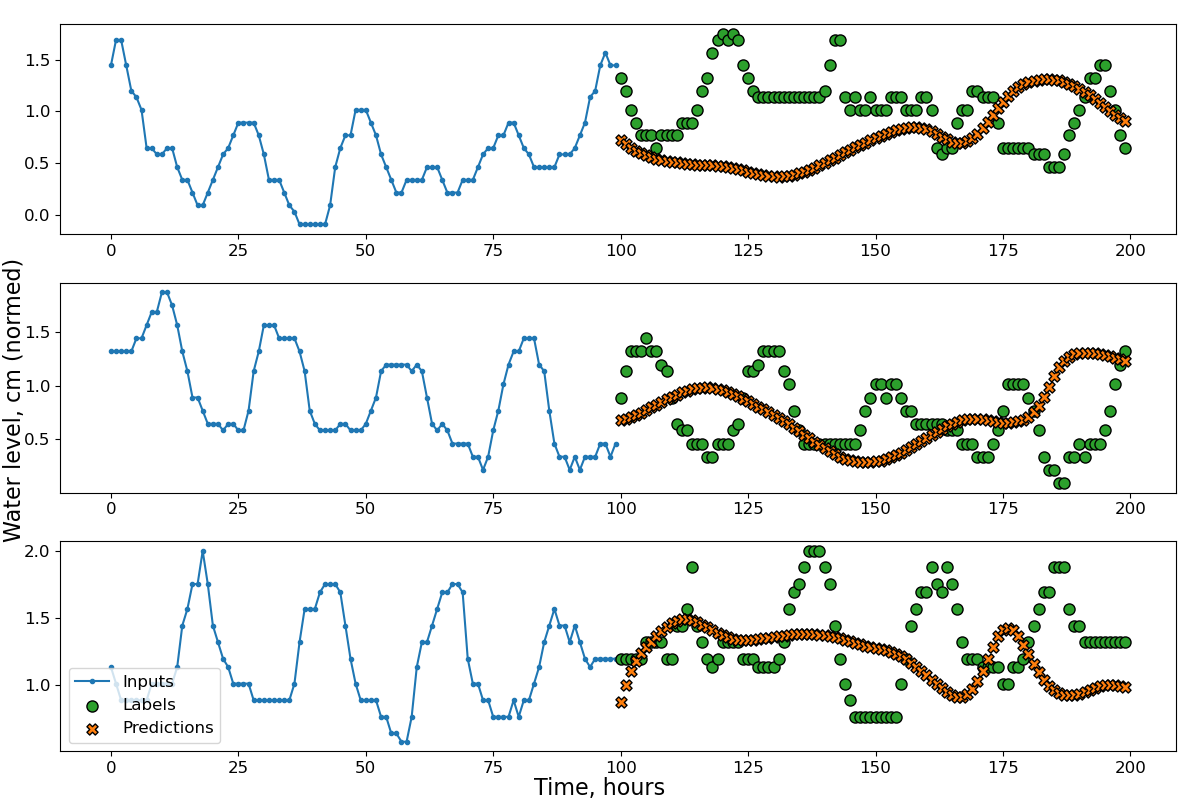}
            \caption{Picked predictions of latent ODE model.}
            \label{fig:latent-ode-output-temp}
        \end{figure}

    For each model, the example of predictions is shown in Figures~\ref{fig:temperature_picked_prediction} and \ref{fig:latent-ode-output-temp}.

    The model prediction ability decreases as the prediction window and noise level increase in most cases. While giving near-constant prediction, a fully-connected model has a significantly lower error than the repeater. We see in Figure~\ref{fig:temperature_picked_prediction_dense} that this model has learned the fact that the time series is periodic with the period of $24$ hours but extrapolates the later $100$ hours with the value close to the last observed point, which becomes less reasonable with the growth of the prediction window.
    
    The seasonal ARIMA model showed a reasonable approximation since the periodic behaviour of the data can be modelled using the previous observations with a fixed time shift. This model shows the second performance after the LSTM model, and, with the $250$ data points to predict, it approximates the data better than the LSTM model.
    
    LSTM model predictions also may be interpreted as an approximation of the seasonal component of the time series. From the picked examples, we can see that this is crucial for the good quality of the predictions. However, as its predictions are stationary, with the growth of the prediction window, the global trend of a time series becomes more significant, and local approximation becomes insufficient, as we can see from decreasing performance for larger window lengths.
    The resulting MAE distribution computed for every window prediction in the dataset is shown in Table~\ref{methods-comarison-temp}.
    More significant preparation was required to experiment with the latent ODE model. Specifically, it was required
    to adapt the implementation from~\cite{latent-ode-for-time-series} for fitting on arbitrary data.
        
        \begin{table}[ht!]
         \caption{Comparison of temperature dataset MAE for different algorithms.}
            \centering
            \small
            \begin{tabular}{cccccc}
               \hline
                \toprule
                \multirow{2}{*}{Algorithm} & \multirow{2}{*}{Noise level} & \multicolumn{3}{c}{MAE} \\
                & & 100 points                 & 250 points                 & 500 points                 \\ \midrule
                \multirow{4}{*}{Repeater}   & $0$ & $0.505 \pm 0.000$          & $0.616 \pm 0.000$          & $0.548 \pm 0.000$        \\ 
                   & $0.1$ & $0.520 \pm 0.000$          & $0.626 \pm 0.000$          & $0.560 \pm 0.000$        \\ 
                   & $0.2$ & $0.549 \pm 0.000$          & $0.647 \pm 0.000$          & $0.594 \pm 0.000$        \\ 
                   & $0.3$ & $0.613 \pm 0.000$          & $0.697 \pm 0.000$          & $0.645 \pm 0.000$        \\ \midrule
                \multirow{4}{*}{FCNN}   & $0$ & $0.432 \pm 0.002$          & $0.501 \pm 0.002$          & $0.487 \pm 0.005$        \\ 
                   & $0.1$ & $0.453 \pm 0.002$          & $0.514 \pm 0.002$          & $0.489 \pm 0.002$        \\ 
                   & $0.2$ & $0.501 \pm 0.003$          & $0.513 \pm 0.001$          & $0.559 \pm 0.004$        \\ 
                   & $0.3$ & $0.509 \pm 0.001$          & $0.586 \pm 0.001$          & $0.627 \pm 0.003$        \\ \midrule
                \multirow{4}{*}{ARIMA}   & $0$ & $0.390 \pm 0.000$          & $\mathbf{0.407 \pm 0.000}$         & $0.460 \pm 0.000$        \\ 
                   & $0.1$ & $0.407 \pm 0.000$          & $\mathbf{0.415 \pm 0.000}$          & $0.469 \pm 0.000$        \\ 
                   & $0.2$ & $0.428 \pm 0.000$          & $\mathbf{0.442 \pm 0.000}$          & $0.492 \pm 0.000$        \\ 
                   & $0.3$ & $0.470 \pm 0.000$          & $\mathbf{0.485 \pm 0.000}$          & $0.531 \pm 0.000$        \\ \midrule
                \multirow{4}{*}{LSTM}   & $0$ & $\mathbf{0.370 \pm 0.015}$          & $0.418 \pm 0.005$          & $\mathbf{0.442 \pm 0.007}$        \\ 
                   & $0.1$ & $\mathbf{0.386 \pm 0.006}$          & $0.426 \pm 0.007$          & $\mathbf{0.449 \pm 0.012}$        \\ 
                   & $0.2$ & $\mathbf{0.410 \pm 0.009}$          & $0.446 \pm 0.007$          & $\mathbf{0.481 \pm 0.013}$        \\ 
                   & $0.3$ & $\mathbf{0.458 \pm 0.007}$          & $0.502 \pm 0.008$          & $\mathbf{0.527 \pm 0.017}$        \\ \midrule
                \multirow{4}{*}{Latent ODE}   & $0$ & $0.500 \pm  0.030$          & $0.496 \pm 0.010$          & $0.515 \pm 0.018$        \\ 
                   & $0.1$ & $0.509 \pm 0.019$          & $0.499 \pm 0.023$          & $0.554 \pm 0.039$        \\ 
                   & $0.2$ & $0.540 \pm 0.007$          & $0.545 \pm 0.025$          & $0.571 \pm 0.022$        \\ 
                   & $0.3$ & $0.576 \pm 0.027$          & $0.557 \pm 0.012$          & $0.607 \pm 0.016$        \\ \bottomrule \hline
            \end{tabular}
            \label{methods-comarison-temp}
        \end{table}

    However, despite many iterations, while the latent ODE model achieved better performance than the repeater model, it was unable to reproduce the periodic components of the data, and its predictions are too smooth and not reasonable for considered time series.
    From the predictions of the best-performing ARIMA and LSTM models, we can see that their key property was proper modelling
    of the periodic components of the data.
    We address the theoretical issues regarding training periodic and other functions with neural ODEs in Section~\ref{sec:node_analysis}.
    
    It should be emphasized that in~\cite{latent-ode-for-time-series} authors achieved better results than classical RNN models on irregularly-sampled data using ODE networks, also showing that when the data is close to the regularly sampled, these models behave as good as classical ones. While this model should easily represent the periodic functions, as expected from the solution of an ODE system, predictions were not periodic in our experiments. One possible reason for this is that in~\cite{latent-ode-for-time-series} authors used ODE parametrization of higher-order, used an encoder network with more parameters, or trained the network for a more extended period for better prediction quality. All these factors make practical use of neural ODE less feasible.
    
    \subsection{Experiments with water level dataset}
    \label{sec:water-level-experiment}

\begin{figure}[ht!]
            \centering
            \includegraphics[width=0.5\linewidth]{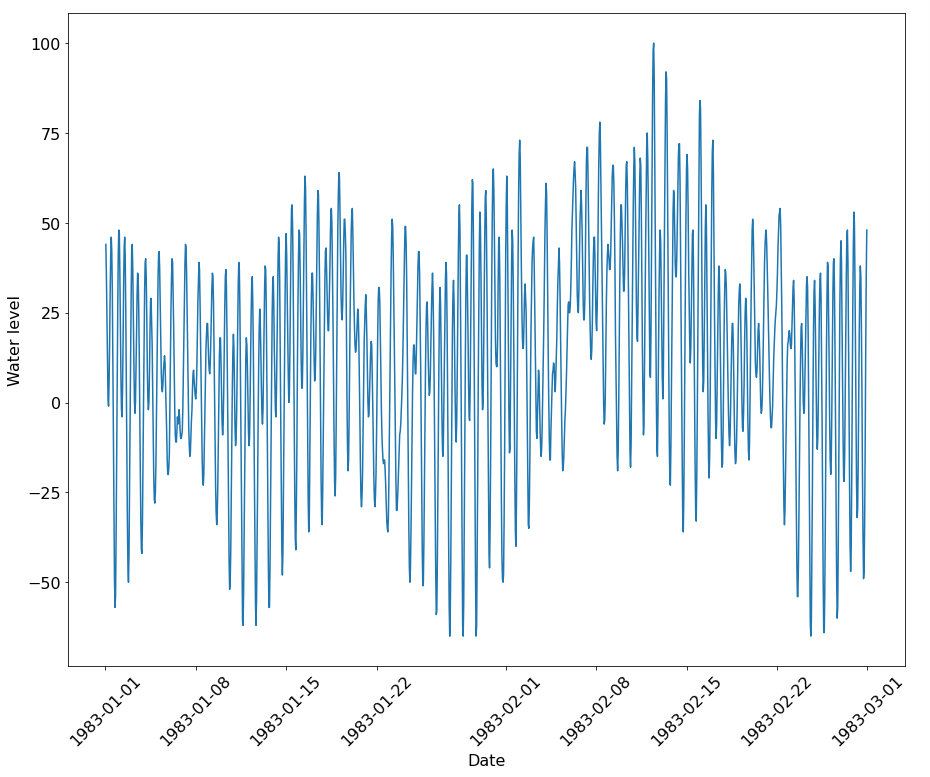}
            \caption{Water level for three months in Venice, Italy.}
            \label{data-example-water-level}
        \end{figure}

\begin{figure*}
        \centering
        \begin{subfigure}[b]{0.495\textwidth}
            \centering
            \includegraphics[width=\textwidth]{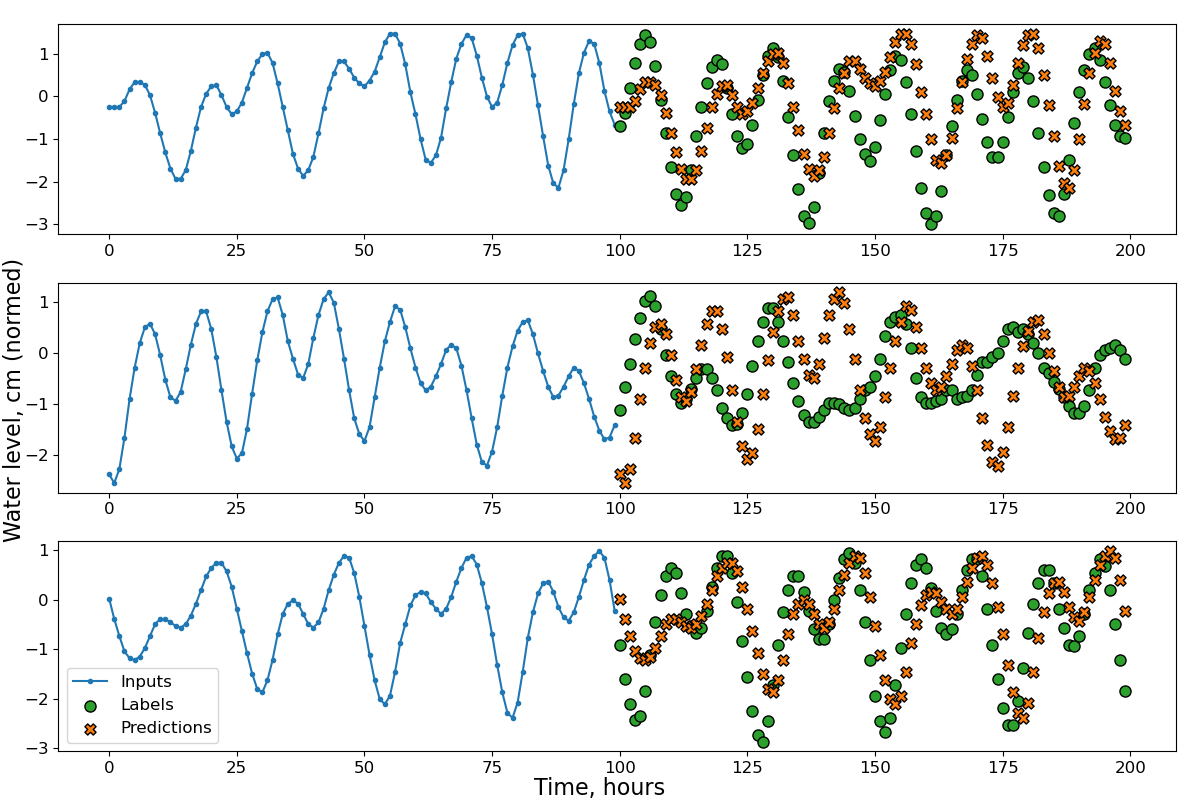}
            \caption[Network2]%
            {\small{Repeater model}}    
            \label{fig:water_level_picked_prediction_repeater}
        \end{subfigure}
        \hfill
        \begin{subfigure}[b]{0.495\textwidth}  
            \centering 
            \includegraphics[width=\textwidth]{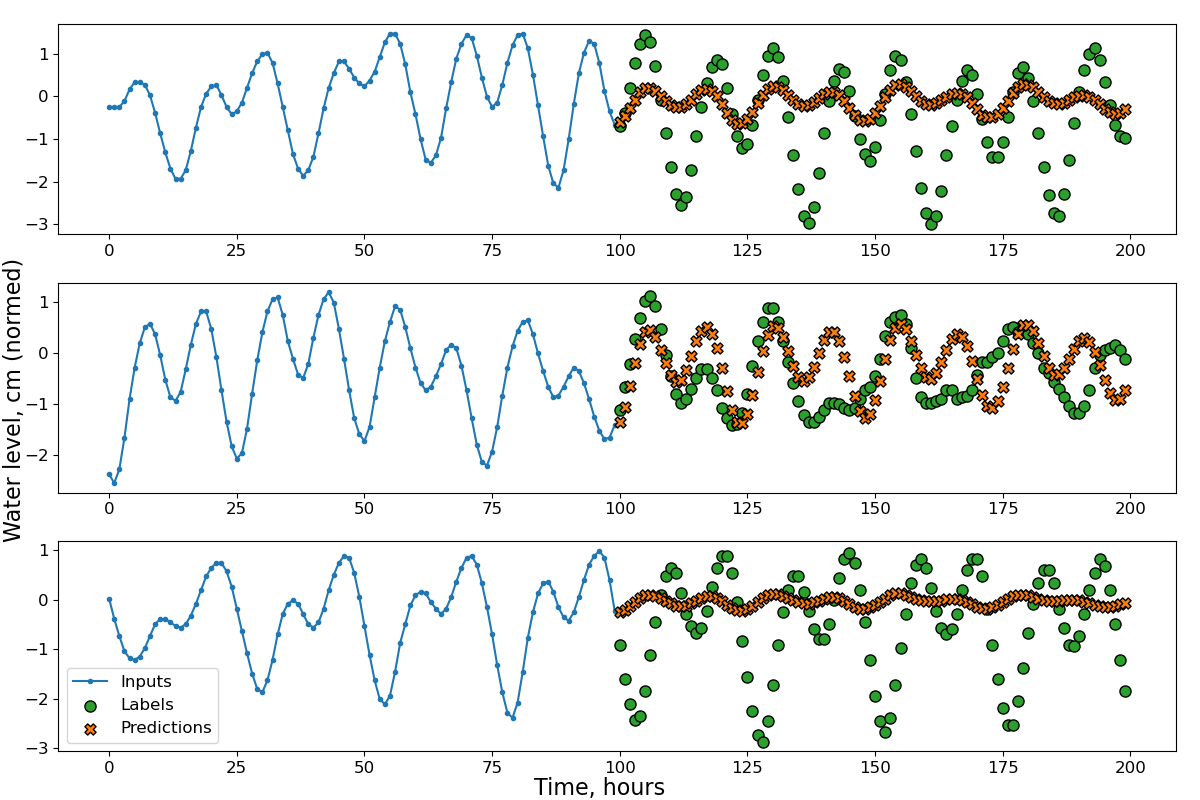}
            \caption[]%
            {\small{Dense neural network}}
            \label{fig:water_level_picked_prediction_dense}
        \end{subfigure}
        \begin{subfigure}[b]{0.495\textwidth}   
            \centering 
            \includegraphics[width=\textwidth]{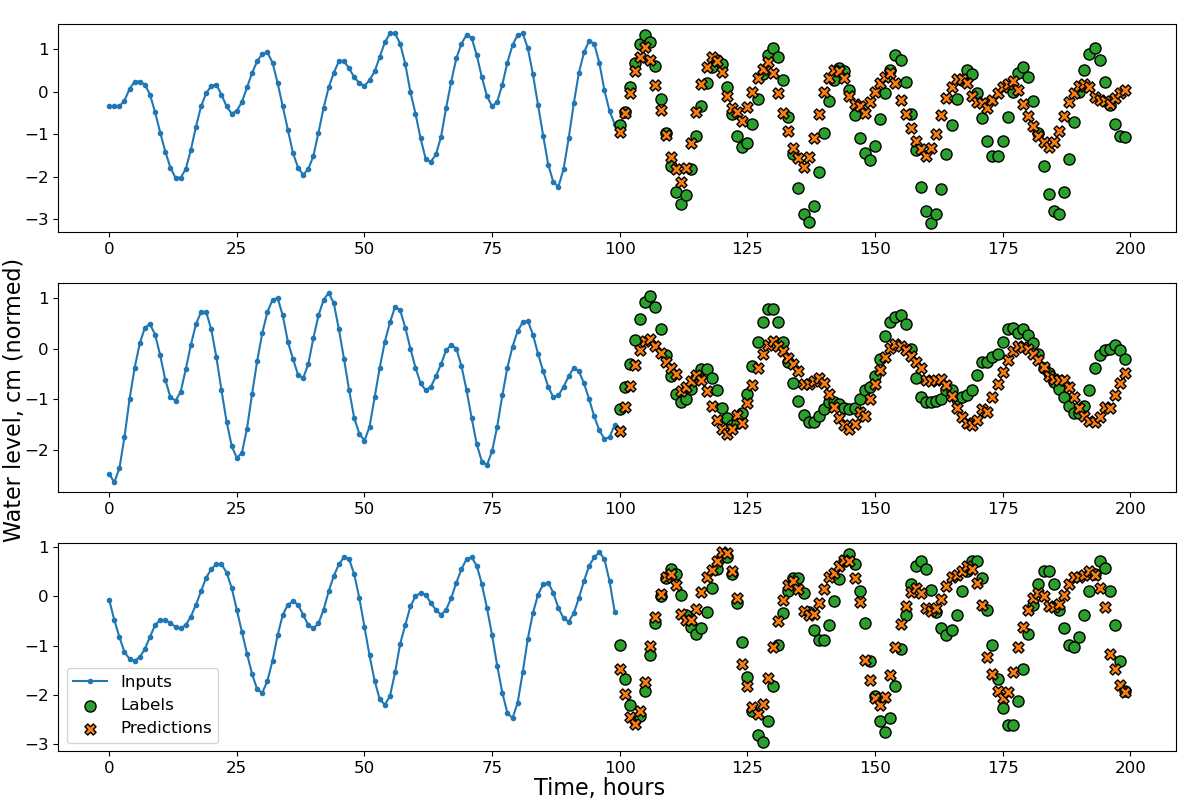}
            \caption[]%
            {\small{Autoregressive model}}
            \label{fig:water_level_picked_prediction_autoreg}
        \end{subfigure}
        \hfill
        \begin{subfigure}[b]{0.495\textwidth}   
            \centering 
            \includegraphics[width=\textwidth]{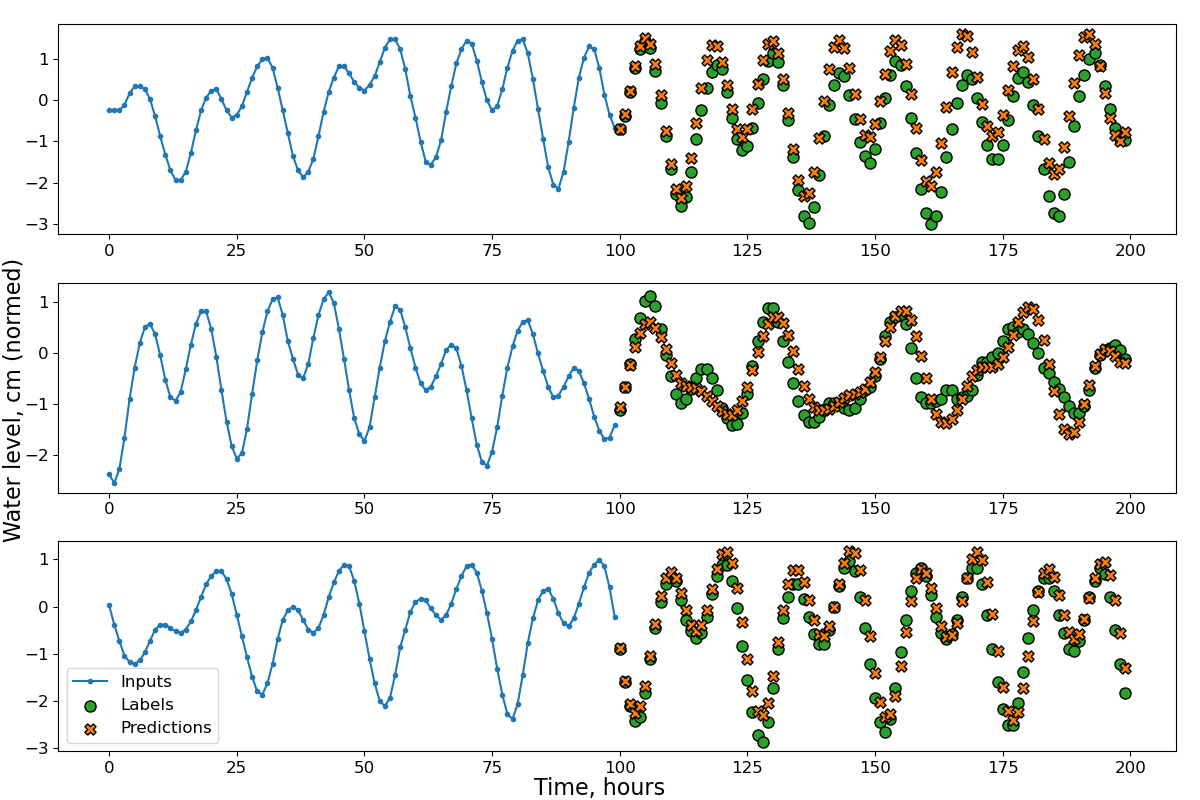}
            \caption[]%
            {\small{LSTM model}}
            \label{fig:water_level_picked_prediction_lstm}
        \end{subfigure}
        \caption[ Picked predictions of a water level dataset for different models. ]
        {\small Picked predictions of a water level dataset for different models.} 
        \label{fig:water_level_picked_prediction}
    \end{figure*}

\begin{figure}[ht!]
            \centering
            \includegraphics[width=0.5\linewidth]{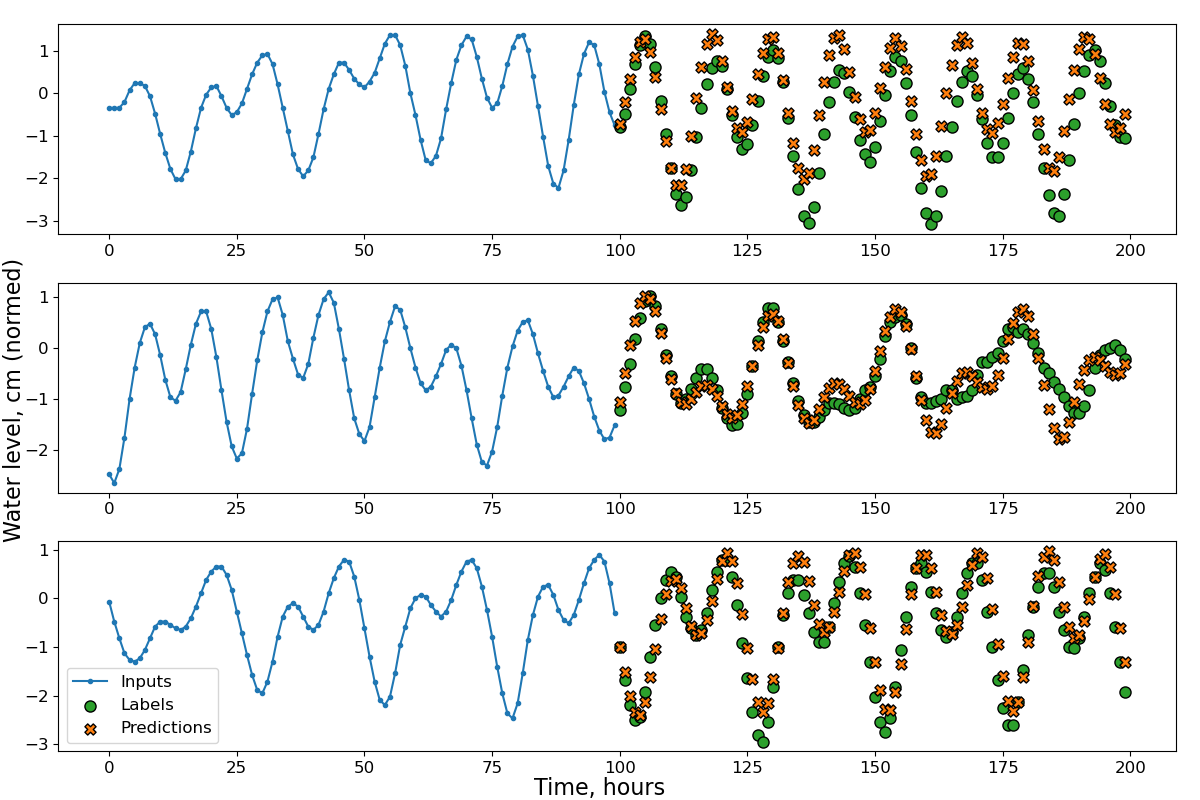}
            \caption{Picked predictions of latent ODE model.}
            \label{latent-ode-output-water-level}
        \end{figure}

    In the second experiment, we selected the same setup as in the previous Section~\ref{sec:temp_experiment}.
    The example of the data for three months is shown in Figure~\ref{data-example-water-level}.
    Similarly, example predictions for each model are shown in Figures~\ref{fig:water_level_picked_prediction}--\ref{latent-ode-output-water-level}.
    
    For this dataset, the repeater model showed more reasonable predictions than for the previous experiment since the behaviour of the considered time series for several successive days is almost the same.
    
    The fully-connected model was able to approximate the periodic behaviour of the data slightly while still failing to model its phase and amplitude properly. This causes it to produce low-quality predictions comparable to the repeater model.
    
    The ARIMA model also showed the good quality of the predictions for the same reason as in the previous experiment, while this time, latent ODE has a lower value of the MAE metric.
    
    LSTM models showed the best prediction quality and the lowest metric value, achieving the best overall result and almost perfect forecasting for picked samples while having significantly worse quality for the window of the length of $500$.
    The resulting MAE values computed for every prediction window and noise level on the water level dataset are shown in Table~\ref{methods-comarison-water-level}.
        
        \begin{table}[ht!]
            \centering
            \small
        \caption{Comparison of water level dataset MAE for different algorithms.}
            \begin{tabular}{cccccc}
                \hline
                \toprule
                \multirow{2}{*}{Algorithm} & \multirow{2}{*}{Noise level} & \multicolumn{3}{c}{MAE} \\
                & & 100 points                 & 250 points                 & 500 points                 \\ \midrule
                \multirow{4}{*}{Repeater}   & $0$ & $0.759 \pm 0.000$          & $0.968 \pm 0.000$          & $1.062 \pm 0.000$        \\ 
                   & $0.1$ & $0.768 \pm 0.000$          & $0.974 \pm 0.000$          & $1.069 \pm 0.000$        \\ 
                   & $0.2$ & $0.787 \pm 0.000$          & $0.990 \pm 0.000$          & $1.086 \pm 0.000$        \\ 
                   & $0.3$ & $0.829 \pm 0.000$          & $1.017 \pm 0.000$          & $1.113 \pm 0.000$        \\ \midrule
                \multirow{4}{*}{FCNN}   & $0$ & $0.761 \pm 0.001$          & $0.803 \pm 0.003$          & $0.932 \pm 0.001$        \\ 
                   & $0.1$ & $0.767 \pm 0.001$          & $0.807 \pm 0.001$          & $0.939 \pm 0.002$        \\ 
                   & $0.2$ & $0.784 \pm 0.002$          & $0.817 \pm 0.002$          & $0.943 \pm 0.002$        \\ 
                   & $0.3$ & $0.805 \pm 0.001$          & $0.844 \pm 0.001$          & $0.964 \pm 0.001$        \\ \midrule
                \multirow{4}{*}{ARIMA}   & $0$ & $0.563 \pm 0.000$          & $0.731 \pm 0.000$          & $\mathbf{0.764 \pm 0.000}$        \\ 
                   & $0.1$ & $0.582 \pm 0.000$          & $0.733 \pm 0.000$          & $\mathbf{0.774 \pm 0.000}$        \\ 
                   & $0.2$ & $0.614 \pm 0.000$          & $0.749 \pm 0.000$          & $\mathbf{0.789 \pm 0.000}$        \\ 
                   & $0.3$ & $0.649 \pm 0.000$          & $0.777 \pm 0.000$          & $\mathbf{0.797 \pm 0.000}$        \\ \midrule
                \multirow{4}{*}{LSTM}   & $0$ & $\mathbf{0.359 \pm 0.008}$          & $\mathbf{0.508 \pm 0.013}$          & $0.803 \pm 0.015$        \\ 
                   & $0.1$ & $\mathbf{0.369 \pm 0.003}$          & $\mathbf{0.519 \pm 0.006}$          & $0.811 \pm 0.015$        \\ 
                   & $0.2$ & $\mathbf{0.407 \pm 0.005}$          & $\mathbf{0.544 \pm 0.007}$          & $0.815 \pm 0.012$        \\ 
                   & $0.3$ & $\mathbf{0.452 \pm 0.003}$          & $\mathbf{0.592 \pm 0.013}$          & $0.844 \pm 0.012$        \\ \midrule
                \multirow{4}{*}{Latent ODE}   & $0$ & $0.469 \pm 0.069$          & $0.787 \pm 0.008$          & $0.855 \pm 0.028$        \\ 
                   & $0.1$ & $0.526 \pm 0.101$          & $0.802 \pm 0.013$          & $0.857 \pm 0.017$        \\ 
                   & $0.2$ & $0.555 \pm 0.100$          & $0.810 \pm 0.010$          & $0.856 \pm 0.005$        \\ 
                   & $0.3$ & $0.581 \pm 0.112$          & $0.839 \pm 0.017$          & $0.889 \pm 0.017$        \\ \bottomrule \hline
            \end{tabular}
            
            \label{methods-comarison-water-level}
        \end{table}

In this experiment, the latent ODE model was able to spot the periodicity of the signal, as shown in Figure~\ref{latent-ode-output-water-level}.
However, in most of the $k$ experiments, it still failed to model both periodic components properly, as suggested by far the largest variance value.
Apart from extremely slow fitting, when increasing the prediction window, this model could not reproduce the periodic behaviour achieved on a window of length $n = 100$, so the value of the metrics increased drastically for the larger windows.

\subsection{When we should prefer neural ODE in time series modelling}

There are many complexities regarding the neural ODE training process, including its inability to converge to the function that approximates the target time series perfectly and extremely slow convergence compared to standard neural networks. We mark the following points for a practical time-series prediction applications:

\begin{itemize}
    \item To achieve superior results, it is required to enormously extend the number of parameters (complexity) of the neural ODE model extending the training and inference time. That significantly reduces the interpretability and stability of the training process overall.
    \item The more complex the system is, the more complex the eigenvalues and eigenvectors space. Every numerical procedure in this case, including training and eigenvalues computation, becomes unstable, complicating the interpretation process.
    \item As a result, apart from the impossibility of decomposing neural ODE, we meet the impossibility of working with the neural ODE as a system of non-linear ODE due to its size.
\end{itemize}

Thus the \textbf{answer} should be that the neural ODE can be applied in a very few modelling cases, as an example in the irregular time-grid as considered in \cite{liquid-time-constant-networks} or where the data has too high noise level. Neural ODEs have no significantly different interpretation abilities from neural networks. Another approach is to obtain closed-form equation as it is done within the PINN \cite{raissi2019physics} or equation discovery frameworks \cite{brunton2016discovering, maslyaev2021partial}, which we will discuss in Section~\ref{sec:discussion}.

\section{Linear neural ODE interpretation in the scope of classical ODE analysis}
\label{sec:node_analysis}

After we understand that we cannot interpret neural ODE as a non-linear system due to its complexity, we could try to make the process simpler. The most straightforward interpretable ODE system that can be obtained with machine learning methods is the single-layer ODE system without the activation function between layers, equivalent to the linear ODE system. Despite its simplicity, it still can be used as a tool for the problem \enquote{linearization} and further spectral analysis.

    
Since the model consists of multiple components, including an encoder network that converts the input signal into initial conditions and an ODE network that performs integration of a hidden state using these initials conditions, such analysis is challenging to perform directly. While implementation from~\cite{latent-ode-for-time-series} uses several fully connected layers with non-linearities to model the hidden state, we consider the case when the hidden state dynamics is represented by a system of linear ODEs and the initial condition are fixed, as many time series can be approximated well using this parametrization, resulting in more transparent analysis.

\subsection{Simple periodic function approximation using system representation}

The goal of the first experiment was to check how well the model was able to learn the periodic functions. We note that the same system could easily be obtained with classic symbolic regression methods.
In order to model a system of linear ODEs, we built a neural ODE that has the ODE network defined by a single fully-connected layer without activation and set up the initial conditions manually for the considered dataset.
In this experiment, fit this model to reproduce function:

    \begin{equation}
        x(t)=\begin{bmatrix}
            \sin(t) \\
            \cos(t)
        \end{bmatrix}
    \label{eq:sine_function_part_sln}
    \end{equation}
    in the range $t \in [0, 2 \pi]$.
    This function is a particular solution of the following ODE:
    \begin{equation}
        \left\{
        \begin{array}{l}
            \frac{d x_1}{ d t} = x_2(t) \\
            \frac{d x_2}{ d t} = -x_1(t)
        \end{array}
        \right.
    \end{equation}
    
    To train the corresponding neural ODE, we uniformly sample $100$ values from the target function~\eqref{eq:sine_function_part_sln} at interval $[0, 2 \pi]$. Up to $1000$ training iterations were required to achieve the relatively close approximation, depending on the initial approximation. As a result, the network parameters correspond to the following ODE:
    \begin{equation}
        \left\{
        \begin{array}{l}
            \frac{d x_1}{ d t} =  0.0039 x_1(t)  + 0.9998 x_2(t) \\
            \frac{d x_2}{ d t} = -0.9997 x_1(t) - 0.0040 x_2(t)
        \end{array}
        \right.
    \end{equation}

This experiment demonstrates the way considered class of models can learn the periodic functions due to its ODE nature, which is crucial for modelling various time series, as shown in experiments in Section~\ref{sec:ML_compare}.
We also note that for a single-frequency trigonometric pair, as in this case, matrix optimization in space $\mathbb{R}^{2 \times 2}$ is required.

Extending this experiment to systems of a larger order can prove that a neural ODE of this form could approximate every continuous time series.
This follows from the two facts: a converging Fourier series can represent every continuous function with compact support.
Furthermore, a finite number of terms of every converging Fourier series can be exactly represented by a solution of the linear ODE.
The second fact follows from the general form of the linear ODE solution, shown in~\eqref{eqn:general-solution} --- 
eigendecomposition allows constructing a matrix with a given set of eigenvalues and eigenvectors that represents the amplitudes 
and frequencies of a given Fourier series terms.
    
\subsection{General view on neural ODE represented by linear system}

A more general view of a training process of the simplest one-layer linear neural ODE for time series modelling is that it consists of three separate models having their own set of parameters that can be estimated using a backpropagation algorithm in a single training procedure:

    \begin{enumerate}
        \item \label{itm:first} Coefficients of the linear system of ODEs, which defines the class of functions that can be represented using this neural network.
        \item Parameters of a neural network that converts the input data into initial conditions for integrating the system of ODEs.
        \item \label{itm:third} Parameters of a neural network that combines the vector of ODE solutions into the target scalar function.
    \end{enumerate}

Note that these models correspond to steps 2-4 of neural ODE workflow in Figure~\ref{fig:neural_ODE_scheme}.
From an interpretability point of view, model~\ref{itm:first}, which contains the system of equations, corresponds to the \enquote{physical} nature of a time series, and its parameters are estimated during the training process. Thus the governing law of time series is determined by the system is not changing in time.

\begin{figure}[h!]
    \centering
    \includegraphics[width=0.3\linewidth]{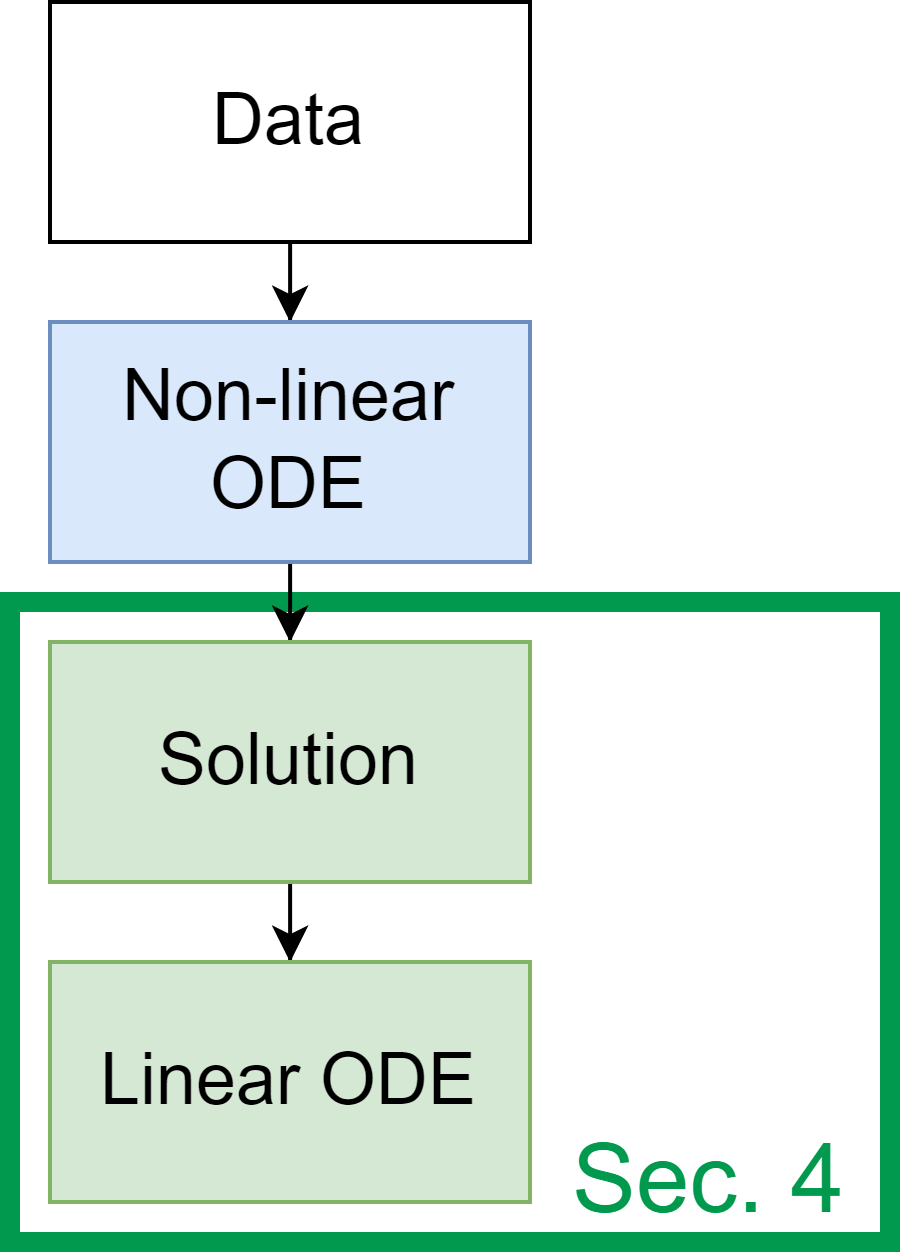}
    \caption{Complete neural ODE linearization scheme.}
    \label{fig:NODE_linearization}
\end{figure}

The whole linearization pipeline is shown in Figure~\ref{fig:NODE_linearization}.

The previous steps are considered in Section~\ref{sec:ML_compare}. In this section, we will discuss only the linearized part.

\paragraph{Numerical example.} As the next step, we consider the following experiment --- given the periodic function with two components, we trained a neural ODE that represents it. The model was trained to minimize the distance between the predictions and the following function:
    \begin{equation}
        x(t) = 4 \sin(t) -5 \sin(2 t)
    \end{equation}
with $100$ points sampled uniformly from $[0, 4 \pi]$.
As a result, obtained neural ODE has the following coefficients in its system of equations:

    \begin{equation}
        \label{eq:optimized-system}
        A = 
        \begin{pmatrix}
            -0.36 & -1.36 & -0.04 & 0.45  \\
            2.36  & 0.43  & -0.47  & -1.06 \\
            0.95  & -0.19 & -0.03  & 0.74  \\
            -0.16 & 1.09  & -1.06  & -0.04
        \end{pmatrix}
    \end{equation}

 Matrix~\eqref{eq:optimized-system} has eigenvalues $\lambda_{1,2} = \pm 2 i, \ \lambda_{3,4} = \pm i$. From eigenvalue analysis we obtain following form of the resulting time series after applying the model~\ref{itm:third} in the list mentioned above:
    \begin{equation}
    \label{eq:node_general_form}
        y(t) = f(c_1 \cos(2 t), c_2 \sin(2 t),c_{3} \cos(x), c_4 \sin(x)) 
    \end{equation}
    
In \eqref{eq:node_general_form} coefficients $c_i$ are determined by model~\ref{itm:first} using initial conditions for ODE system and function $f$ determined by neural network~\ref{itm:third}. In this experiment we used linear function $f$ in~\eqref{eq:node_general_form}. While we perfectly fit the data using this model, the problem is that, similarly to the previous experiment, it took several hundreds of iterations to converge.

This experiment brings us the idea that a system~\eqref{eq:optimized-system} can be optimized implicitly using a set of eigenvalues, opposing to original neural ODE approach.

We want to predict the time series behaviour based on its model for practical tasks. Hence, given the fixed system of ODEs, we need to estimate the initial conditions based on the data prehistory. It turns out that this estimation is nearly impossible for some of the systems using standard neural networks, such as LSTM, as suggested by the following example with the same spectrum as~\eqref{eq:optimized-system} --- given the system of differential equations:

    \begin{equation}
        \label{eqn:bad-equation}
        \begin{cases}
            x_1'(t) = x_2(t) \\
            x_2'(t) = x_3(t) \\
            x_3'(t) = x_4(t) \\
            x_4'(t) = -4 x_1(t) - 5 x_3(t) \\
        \end{cases}
    \end{equation}
    
which corresponds to the following single differential equation:
    \begin{equation}
        x_1(t)'''' + 5 x_1(t)'' + 4 x_1(t) = 0
    \end{equation}
    
The solution is fully determined by the first function $x_1$:
    \begin{equation}
        x_1(t) = c_2 \sin(2 t) + c_4 \sin(t) + c_1 \cos(2 t) + c_3 \cos(t)
    \end{equation}

In case when $y = x_1$, the model has to estimate the first and second derivative of the data to obtain the initial conditions for $x_2$ and $x_3$. For the dense neural networks to represent the function together with its derivatives (the usual neural network training process allows only convergence to a function, but not to its derivative), Sobolev learning process \cite{czarnecki2017sobolev} is required. The question of the existence and application of the Sobolev training process to LSTM networks is still open. Moreover, in this case, the input LSTM has to predict the next value of a time series itself, as it coincides with the initial conditions for $x_1$. If we have such LSTM, we do not need the neural ODE.

\paragraph{Limitations and drawbacks.} While the equation~\eqref{eqn:bad-equation} is an extreme case, and we can not guarantee which system exactly will be obtained during the training process. Moreover, the optimization in the space of ODE systems matrices may be considered an ill-posed problem since the correspondence between a particular solution and the ODE system is non-unique in most cases --- infinitely many systems correspond to the same vector of particular solutions, as there are infinitely many matrices with the same set of eigenvalues.
Regularization may overcome non-uniqueness, but it will also increase the optimization time.
For example, LASSO regularization can be used to obtain a matrix as sparse as possible.

Optimization speed is crucial for this problem. The trivial ODE system case from the last experiment required many iterations to achieve a reasonable approximation of the target functions. Therefore for the complex time series, the optimization of the system of ODEs becomes a computationally complex problem even with the fast neural network regression methods.

While non-uniqueness of the solution is a common property of neural network optimization problems and can sometimes contribute to the ease of finding the solution, the most significant problem related to the neural ODE training is the stability of the system of equations, which hardens by the ill-posedness of the problem overall. For some systems, the change of a single coefficient can drastically affect the form of a solution by adding several exponential terms that will raise the value of a loss function. This problem is well-known and is studied by a separate branch of mathematics --- the stability theory of ODE systems. As a result, an optimization algorithm can leave a good local minimum during the training process and turn in a completely wrong direction due to a single wrong step.

The intuition behind these issues is that the eigenvalues, which define the form of the solution of neural ODE, depend on the systems matrix elements in a complex way.
Eigenvalues, being roots of the characteristic polynomial, can be found explicitly as functions of polynomial coefficients,
while polynomial coefficients themselves are functions of matrix elements.
However, these functions become highly complex as the order of a system of equations and, consequently, characteristic polynomial degree increases.
In particular, it is a well-known result that for general polynomial equations of order five and higher, there is no solution in radicals, which implies that for most of the characteristic polynomials, there is no simple dependence between eigenvalues and matrix
coefficients.
We also note that there exist explicit formulas for the derivatives of eigenvalues with respect to matrix elements~\cite{magnus1985eigenvalues}. However, these formulas are inapplicable in a standard neural ODE training framework.

Undoubtedly, the \enquote{linearization} of the system is \textbf{possible} following neural ODE guidelines. However, its ambiguity and computation speed cast doubt on if this tool should be used instead of classical ones. Therefore, if we aim to obtain an ODE system, not the deep neural network, we should aim for other methods discussed in the last sections.



\subsection{Replacing neural ODE in system form with explicit solution}

The same linearization effect may be achieved when we explicitly use the solution of an ODE in closed form. In case each eigenvalue is simple, the solution of the linear system of ODEs of order $n$ can be written in the following form:
    \begin{equation}
                \label{eqn:general-solution}
        X(t) = \sum_{k=1}^{n / 2}\Big[ C_{2k - 1}\operatorname{Re}\left[e^{(\alpha_k + i \beta_k)t}V_{2k - 1}\right]\  + C_{2k}\operatorname{Im}\left[e^{(\alpha_k + i \beta_k)t}V_{2k}\right] \Big]
    \end{equation}

In~\eqref{eqn:general-solution} $ \alpha_k \pm i \beta_k  =\lambda_k$ are the eigenvalues (every complex eigenvalue should come with corresponding complex conjugate value, thus, the number of values $n$ is always even), $V_k$ are the eigenvectors, and $\operatorname{Im}\left[\lambda_k\right] \neq 0 \ \forall k \in \mathbb{N}$.  Coefficients $C_k$ are implicitly determined by initial conditions, however, we instead can model them explicitly along with the time variable $t$ as the functions $C_k(t)$. This fact gives us an idea that it is possible to fit the solution of linear neural ODE to the data instead of the system of equations.

Instead of fitting $n^2$ coefficients for a linear system of ODEs, we reduce the search space with $n$ parameters for real and imaginary parts of corresponding eigenvalues.

While in~\eqref{eqn:general-solution} there are also present eigenvector multipliers, we can consider only one solution in the vector of solutions, as they differ only in constant multipliers for each term, which can be compensated by coefficients $C_k$. As a result, we can fit the parameters of the following function:
    \begin{equation}
        \label{eqn:simple-solution}
        X(t) = \sum_{k=1}^{n / 2} \Big[ C_{2k-1}e^{\alpha_k t} (\cos(\beta_k t) - \sin(\beta_k t)) + C_{2k}e^{\alpha_k t} (\cos(\beta_k t) + \sin(\beta_k t)) \Big]
    \end{equation}

Each parameter now affects only one component of the solution. In~\eqref{eqn:simple-solution} the vector of coefficients $C_k$ is determined by a separate neural network, similarly to the case of neural ODE. This formulation is a particular case of symbolic regression, which has a drastically reduced problem dimension and makes optimization more straightforward.

To demonstrate the efficiency of this approach, we fit the model~\eqref{eqn:simple-solution} on the water level dataset and compared its performance to the other models from Section~\ref{sec:water-level-experiment}. For simplicity, we set coefficients $\alpha_k$ to $0$, as exponential terms were not required for the dataset.

Given an analytical solution~\eqref{eqn:simple-solution}, we substitute each parameter with obtained value and calculate the required number of time-series observations. Firstly, we fit the coefficients $\beta_k$ and $C_k$ along with the time-shift $t_0$ on a randomly picked sample, obtaining the initial approximation. After that, we fit the neural network that maps the input part of the time series into coefficients $C_k$ and time-shift $t_0$.

\begin{figure}[h!]
            \centering
            \includegraphics[width=0.5\linewidth]{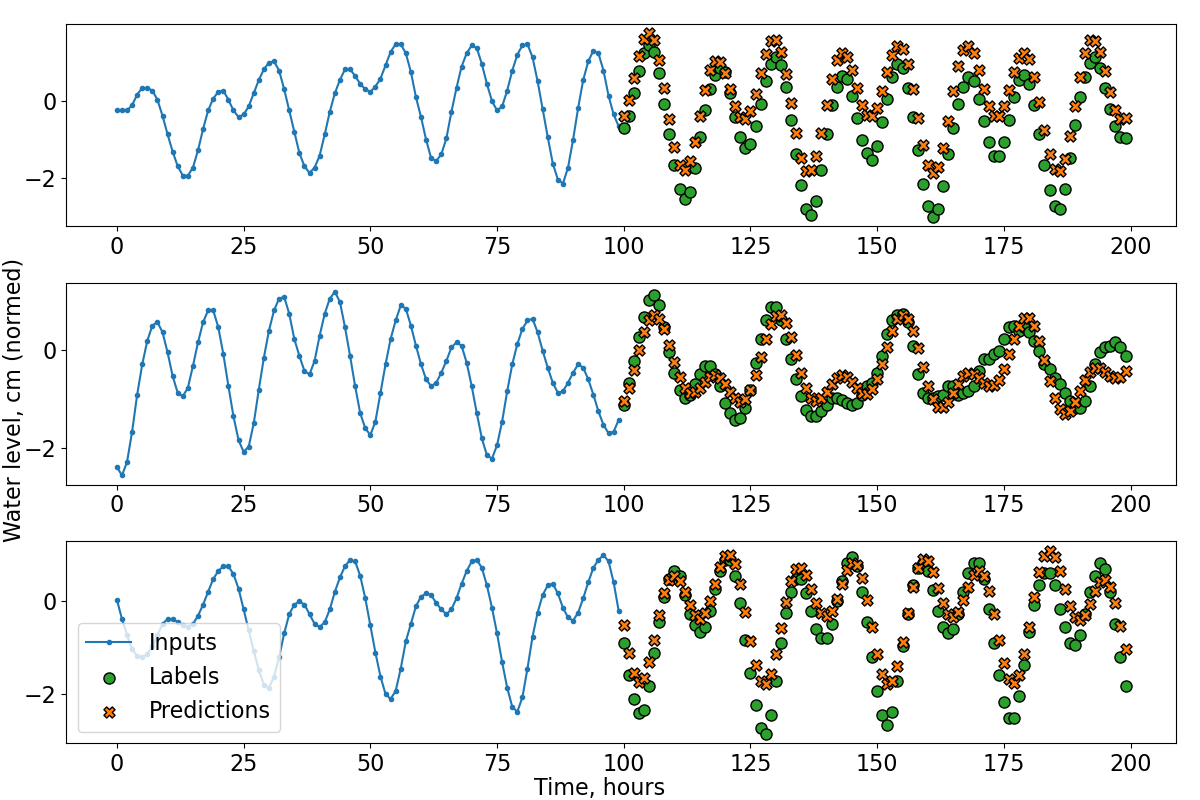}
            \caption{Picked predictions of closed form model.}
            \label{closed-model-output-water-level}
        \end{figure}

In experiments, we used COBYLA~\cite{cobyla} numerical optimization algorithm for initial parameter estimation. Alternatively, it is possible to use a more efficient but restricted approach to estimate the solution parameters using Fourier transform, which gives similar results but can be used only to parameterize a set of periodic functions. Additionally, a neural network can refine the predictions of this model, which further increases the prediction quality but decreases interpretability.
The example predictions for this model are shown in Figure~\ref{closed-model-output-water-level}.
A comparison of the MAE metric on the test dataset for this and several previously discussed models for various window lengths is shown in Table~\ref{closed-form-water-level}.

        \begin{table} [h!]
            \centering
            \small
            \caption{Comparison of water level dataset MAE for different algorithms.}
            \begin{tabular}{cccccc}
               \hline
                \toprule
                \multirow{2}{*}{Algorithm} & \multirow{2}{*}{Noise level} & \multicolumn{3}{c}{MAE} \\
                & & 100 points                 & 250 points                 & 500 points                 \\ \midrule
                \multirow{4}{*}{Closed form}   & $0$ & $0.427 \pm 0.005$          & $0.648 \pm 0.014$          & $\mathbf{0.788 \pm 0.017}$        \\ 
                   & $0.1$ & $0.436 \pm 0.008$          & $0.641 \pm 0.018$          & $0.814 \pm 0.015$        \\ 
                   & $0.2$ & $0.470 \pm 0.004$          & $0.654 \pm 0.011$          & $\mathbf{0.802 \pm 0.023}$        \\ 
                   & $0.3$ & $0.536 \pm 0.011$          & $0.705 \pm 0.024$          & $\mathbf{0.829 \pm 0.030}$        \\ \midrule
                \multirow{4}{*}{LSTM}   & $0$ & $\mathbf{0.359 \pm 0.008}$          & $\mathbf{0.508 \pm 0.013}$          & $0.803 \pm 0.015$        \\ 
                   & $0.1$ & $\mathbf{0.369 \pm 0.003}$          & $\mathbf{0.519 \pm 0.006}$          & $\mathbf{0.811 \pm 0.015}$        \\ 
                   & $0.2$ & $\mathbf{0.407 \pm 0.005}$          & $\mathbf{0.544 \pm 0.007}$          & $0.815 \pm 0.012$        \\ 
                   & $0.3$ & $\mathbf{0.452 \pm 0.003}$          & $\mathbf{0.592 \pm 0.013}$          & $0.844 \pm 0.012$        \\ \midrule
                \multirow{4}{*}{Latent ODE}   & $0$ & $0.469 \pm 0.069$          & $0.787 \pm 0.008$          & $0.855 \pm 0.028$        \\ 
                   & $0.1$ & $0.526 \pm 0.101$          & $0.802 \pm 0.013$          & $0.857 \pm 0.017$        \\ 
                   & $0.2$ & $0.555 \pm 0.100$          & $0.810 \pm 0.010$          & $0.856 \pm 0.005$        \\ 
                   & $0.3$ & $0.581 \pm 0.112$          & $0.839 \pm 0.017$          & $0.889 \pm 0.017$        \\ \bottomrule \hline
            \end{tabular}
            \label{closed-form-water-level}
        \end{table}

Due to a significantly lower number of parameters and more interpretable search space, model~\eqref{eqn:simple-solution} converged much faster than its ODE counterpart. Compared to fitting coefficients of the system of equations directly, while having the same representational ability, this approach is much more stable. We obtained the optimal values of the coefficients in each experiment and did not fall into the local optima.

Model~\eqref{eqn:simple-solution} can be generalized to a broader class of ODEs, not only linear. One significant limitation is that the solution of the equation should be represented as the closed-form expression. When a general form of a neural ODE solution is available, alternative evolutionary optimization-based approaches for symbolic regressions can be applied~\cite{closed-form-algebraic-discovery} to obtain a model of a non-restricted form.

\subsection{What is neural ODE interpretation}

Working with a solution instead of the equation is \textbf{maximal} information that we can extract from neural ODE for the practical time-series application. On the one hand, we can restore the equation to work with the physics of the system. On the other hand, the overall performance increase is the expected behaviour of the optimization process, as it becomes much more straightforward compared to the optimization of the linear ODE system. 

One may try to obtain a more general form of the system's solution. In the linear case, we use equivalent to the extended version of the Fourier transform as shown in ~\eqref{eqn:simple-solution}. In other cases, we must obtain the general form containing special functions or distributions. As a drawback, we should know a priori what the eigenvector form is expected. We also may obtain the compact approximate model without any a priori form using the symbolic regression methods for a closed-form expression.


From the time series models' interpretability point of view, the latent neural ODE does not have many advantages over the classical ML methods. Therefore, we may consider alternatives, such as equation discovery, which allow obtaining differential equations faster. Additionally, it gives a more convenient equation form for the classical analysis. That is discussed in the next section.

\section{Discussion}
\label{sec:discussion}

Previously we noticed that the considered models could show unexpectedly bad results in prediction quality and training performance. By studying the linear neural ODE and replacing it with an explicit solution, we improved the overall model performance, surpassing latent ODE model form~\cite{latent-ode-for-time-series} and showing performance close to LSTM networks for a specific dataset. While the proposed approach is trivial, it reveals essential insights into how the training and the inference of various neural ODE models can be optimized.
    
In fact, this approach is applicable even for more general cases of ODEs, such as liquid time-constant (LTC) networks~\cite{liquid-time-constant-networks,closed-form-continuous-models}. However, this approach is inapplicable when we cannot obtain a closed expression for the solution, which is true for most non-linear ODEs.
    
While there is a broad class of functions that a linear ODE can represent with the closed-form solution, there are functions for which linear representation will be inaccurate and require large order of the system to obtain an acceptable approximation.
    
On the other hand, directly optimizing the coefficients of the system of equations is highly inefficient, even for the most straightforward systems. In the case of more complicated non-linear ODEs, a non-linear numerical solver is required, making this problem even harder.

\begin{figure}[ht!]
    \centering
    \includegraphics[width=0.4\linewidth]{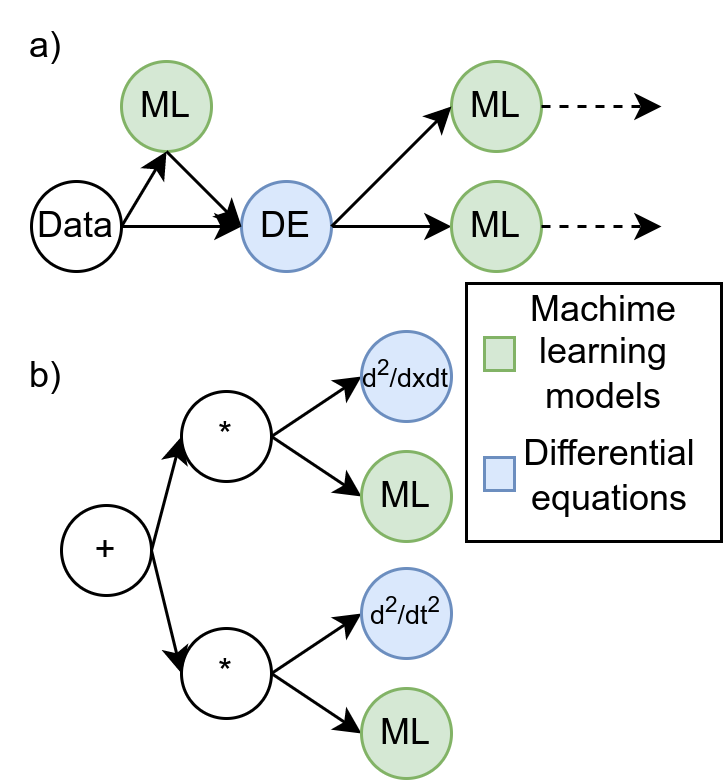}
    \caption{Possible uses of differential equations in the composite machine learning pipelines: a) as a separate model; b) as a model with variable coefficients in form of the machine learning models.}
    \label{fig:node_future}
\end{figure}

Modern automatic machine learning (AutoML) system \cite{nikitin2022automated} allows obtaining various forms of composite models. Another direction of the research may be the differential equation discovery algorithms \cite{maslyaev2021partial} working in the composite machine learning pipelines. The addition of differential equations in such algorithms allows obtaining more interpretable forms of models in two ways shown in Figure~\ref{fig:node_future}.

First is the neural networks used as a parametrization and reduce the error of the given differential equation as shown in Figure~\ref{fig:node_future}a). Second is the neural networks that are used as the variable coefficients of the differential terms as shown in Figure~\ref{fig:node_future}b). Both approaches theoretically should add more interpretation than the existing neural-ODE practices.

\section{Conclusion}
\label{sec:conclusion}

We investigated the modern state of the class of continuous-time neural networks --- neural ODE, with theoretical exploration and experiment conduction from a practical time-series analysis point of view.
    
During the overview part, we observe that lots of the limitations of the original model from~\cite{chen2018neural} are already discovered, many extensions of the model family are proposed, and many practical problems are solved by adapting these models to a specific domain. In our opinion, the process of neural ODE training should not be considered a quality increase but an interpretability increase. The following results are obtained during a theoretical investigation:

\begin{itemize}
    \item Even in the linear ODE system case, the search space is big enough for the complete system optimization, and it takes excessive time to obtain the system's matrix
    \item The linear ODE case may only be interpreted using the spectrum of the matrix. Additionally, we may reduce the regression in the matrix space to the functional one in the space of the (approximate) solutions
    \item While replacing an ODE with its solution might seem trivial for the linear case. This approach gives us insight into how neural ODEs can be effectively trained and applied to the time-series prediction problem.
\end{itemize}

We believe that the future of the neural ODE may not be the solution to the current restrictions of the method, but in creating the machine learning pipelines with differential equations used in machine learning as the equal part of the neural networks with the help of differential equation discovery algorithms.

\bibliographystyle{unsrt}  
\bibliography{references}

\end{document}